\documentclass{article}
\PassOptionsToPackage{numbers, compress}{natbib}
\usepackage[preprint]{neurips_2024}
\usepackage[utf8]{inputenc}
\usepackage[T1]{fontenc}
\usepackage{hyperref}
\usepackage{url}
\usepackage{booktabs}
\usepackage{nicefrac}
\usepackage{microtype}
\usepackage{xcolor}

\usepackage{multicol}
\usepackage{multirow}
\usepackage{pifont} 
\usepackage{amsmath, amssymb, amsfonts}
\usepackage{wrapfig}

\usepackage{algorithm}
\usepackage{graphicx}
\usepackage{threeparttable}
\usepackage{placeins}
\usepackage{subcaption}

\newcommand{\tabincell}[2]{\begin{tabular}{@{}#1@{}}#2\end{tabular}}

\title{Visual Language Tracking with Multi-modal Interaction: A Robust Benchmark}

\author{
\textbf{Xuchen Li}$^{1,2}$ \hspace{9pt}
\textbf{Shiyu Hu}$^{3}$\hspace{9pt} 
\textbf{Xiaokun Feng}$^{1,2}$ \hspace{9pt}
\textbf{Dailing Zhang}$^{1,2}$\hspace{9pt}
\\
\textbf{Meiqi Wu}$^4$ \hspace{9pt}
\textbf{Jing Zhang}$^1$ \hspace{9pt}
\textbf{Kaiqi Huang}$^{1,2,5}$ \hspace{9pt}
\\
$^1$CRISE, Institute of Automation, Chinese Academy of Sciences\\
$^2$School of Artificial Intelligence, University of Chinese Academy of Sciences\\
$^3$School of Physical and Mathematical Sciences, Nanyang Technological University\\
$^4$School of Computer Science and Technology, University of Chinese Academy of Sciences\\
$^5$CAS Center for Excellence in Brain Science and Intelligence Technology\\
lixuchen2024@ia.ac.cn, shiyu.hu@ntu.edu.sg \{fengxiaokun2022, zhangdailing2023\}@ia.ac.cn,\\ wumeiqi18@mails.ucas.ac.cn, \{jing\_zhang, kqhuang\}@ia.ac.cn
}

\begin{document}

\maketitle
\begin{abstract}
Visual Language Tracking (VLT) enhances tracking by mitigating the limitations of relying solely on the visual modality, utilizing high-level semantic information through language. This integration of the language enables more advanced human-machine interaction. The essence of interaction is cognitive alignment, which typically requires multiple information exchanges, especially in the sequential decision-making process of VLT. However, current VLT benchmarks do not account for multi-round interactions during tracking. They provide only an initial text and bounding box (bbox) in the first frame, with no further interaction as tracking progresses, deviating from the original motivation of the VLT task. To address these limitations, we propose a novel and robust benchmark, \textbf{VLT-MI} (\textbf{V}isual \textbf{L}anguage \textbf{T}racking with \textbf{M}ulti-modal \textbf{I}nteraction), which introduces multi-round interaction into the VLT task for the first time. (1) We generate diverse, multi-granularity texts for multi-round, multi-modal interaction based on existing mainstream VLT benchmarks using DTLLM-VLT, leveraging the world knowledge of LLMs. (2) We propose a new VLT interaction paradigm that achieves multi-round interaction through text updates and object recovery. When multiple tracking failures occur, we provide the tracker with more aligned texts and corrected bboxes through interaction, thereby expanding the scope of VLT downstream tasks. (3) We conduct comparative experiments on both traditional VLT benchmarks and VLT-MI, evaluating and analyzing the accuracy and robustness of trackers under the interactive paradigm. This work offers new insights and paradigms for the VLT task, enabling a fine-grained evaluation of multi-modal trackers. We believe this approach can be extended to additional datasets in the future, supporting broader evaluations and comparisons of video-language model capabilities.
\end{abstract}

\section{Introduction}

Single Object Tracking (SOT) has progressed from short-term tracking \cite{otb50,got10k,biodrone} to long-term tracking \cite{lasot}, and more recently to global instance tracking \cite{git}. Researchers are increasingly focusing on making trackers more human-like through interaction and assessing tracking intelligence based on human visual capabilities \cite{git,sotverse}. However, due to the limitations of single-modality systems, SOT faces challenges in intuitively interacting with humans, which impedes its progression toward more complex downstream tasks. Visual Language Tracking (VLT) \cite{otb99,tnl2k} introduces the textual modality, aiming to facilitate more flexible and intuitive human interactions, thus enabling the evaluation of human-like intelligence in trackers. Despite this goal, current VLT trackers \cite{mmtrack,jointnlt,uvltrack,snlt,gti,vlt,transnlt} do not incorporate multi-round, multi-modal interactions during the tracking process. Both training and testing typically provide only a text description and bounding box in the initial frame, which deviates from the original motivation of the VLT task. Moreover, most existing VLT benchmarks \cite{otb99,tnl2k,mgit} annotate texts at a single granularity, focusing mainly on object features in the initial frame. This approach hampers the ability to provide high-quality information through multi-round interactions, leading to misalignment between modalities. At the evaluation level, current benchmark metrics broadly reflect tracker performance but lack fine-grained assessments of robustness.

To address these issues, we develop the VLT-MI benchmark, which allows for dynamic interaction during the tracking process, enabling text updates and object recovery based on the evolving context of the video. We utilize DTLLM-VLT \cite{dtllm} to generate high-quality video textual information for interaction, facilitating improved alignment between visual and textual modalities throughout the tracking process. Additionally, we introduce more detailed, interaction-specific metrics to provide a deeper understanding of how effectively the tracker maintains object continuity and adapts to situational changes, thus offering a more nuanced evaluation of tracker robustness and intelligence.

\section{Construction of VLT-MI}

\begin{table}[t!]
    \vspace{-10pt}
    \centering
    \fontsize{8pt}{10pt}\selectfont
    \setlength{\tabcolsep}{3.5pt}
    \vspace{-8pt}
    \caption{Comparison of current datasets for VLT. VLT-MI is the first VLT benchmark that supports multi-round multi-modal interaction, covering three different tracking tasks. "STT", "LTT" and "GIT" refer to Short-term Tracking, Long-term Tracking and Global Instance Tracking.}
    \label{comparison}
    \begin{threeparttable}
    \begin{tabular}{c|ccccccc}
      \hline
     \multirow{2}{*}{{\tabincell{c}{Dataset}}} & \multirow{2}{*}{{\tabincell{c}{Video\\ number}}} & \multirow{2}{*}{{\tabincell{c}{Min\\ frame}}} & \multirow{2}{*}{{\tabincell{c}{Mean\\ frame}}} & \multirow{2}{*}{{\tabincell{c}{Max\\ frame}}} & \multirow{2}{*}{{\tabincell{c}{Total\\ frames}}} & \multirow{2}{*}{{\tabincell{c}{Tracking\\ task}}} &
     \multirow{2}{*}{{\tabincell{c}{Multi-round\\ Multi-modal Interaction}}} \\ \\
      \hline
\textbf{OTB99\_Lang} & 99    & 71    & 590    & 3,872  & 59K   & STT & $\times$ \\
\textbf{LaSOT}   	 & 1,400 & 1,000 & 2,506  & 11,397 & 3.52M & LTT & $\times$\\
\textbf{TNL2K}       & 2,000 & 21  	 & 622    & 18,488 & 1.24M & STT & $\times$\\
\textbf{MGIT}        & 150   & 4,008 & 14,920 & 29,834 & 2.03M & GIT & $\times$   \\
      \hline
\textbf{VLT-MI}      & 3,619 & 21  	 & 1,824  & 29,834 & 6.60M & STT \& LTT \& GIT & \checkmark   \\
      \hline
    \end{tabular}
    \end{threeparttable}
    \vspace{-20pt}
\end{table}

\begin{wrapfigure}{r}{0.48\textwidth}
\centering
\vspace{-10pt}
\includegraphics[width=0.48\textwidth]{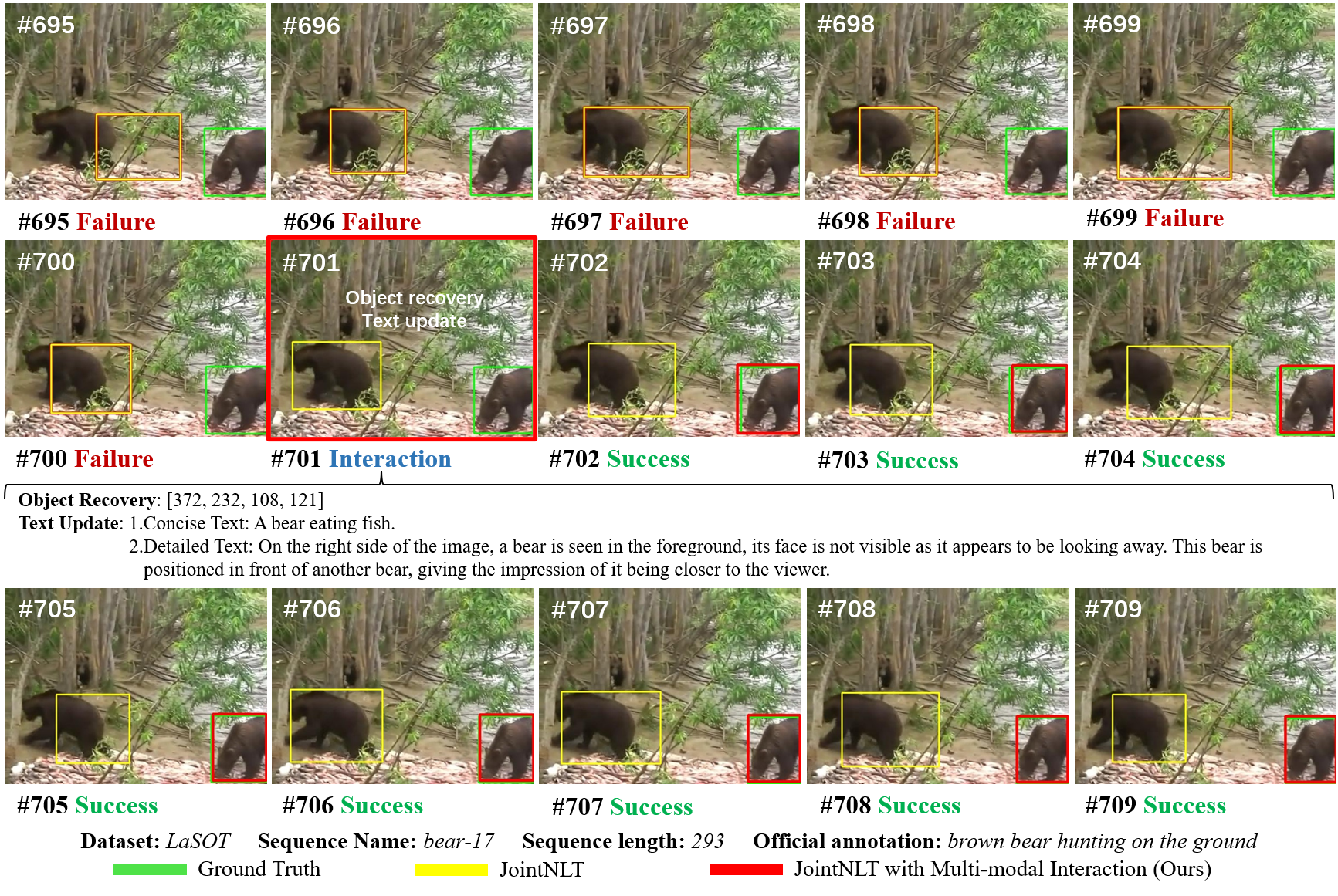}
\vspace{-8pt}
\caption{Example of tracking with multi-round multi-modal interaction. We provide the tracker with the correct bbox and a more accurately aligned concise/detailed text through interaction.}
\label{method}
\vspace{-10pt}
\end{wrapfigure}

We utilize DTLLM-VLT \cite{dtllm} to provide concise and detailed textual descriptions for videos in four mainstream VLT datasets every 100 frames. As shown in Table \ref{comparison}, VLT-MI is the first VLT benchmark with multi-round multi-modal interactions. During the interaction process, if the bbox Intersection over Union (IoU) between the predicted object and the ground truth across 10 consecutive frames falls below 0.5, we consider the tracker to have lost accurate tracking of the object. In such cases, interaction is required to guide the tracker. This interaction is achieved by updating the textual information closest to the interaction frame and providing an accurate bounding box for the current frame. To enhance the quality of textual information during interaction, we generate both concise and detailed descriptions. The concise text focuses solely on the essential features of the object, while the detailed text also incorporates background descriptions. The interaction process is illustrated in Figure \ref{method}.

\section{Experimental Results}

\begin{figure}[ht]
    \vspace{-10pt}
    \centering
    \begin{minipage}{0.48\textwidth}
        \centering
        \includegraphics[width=\textwidth]{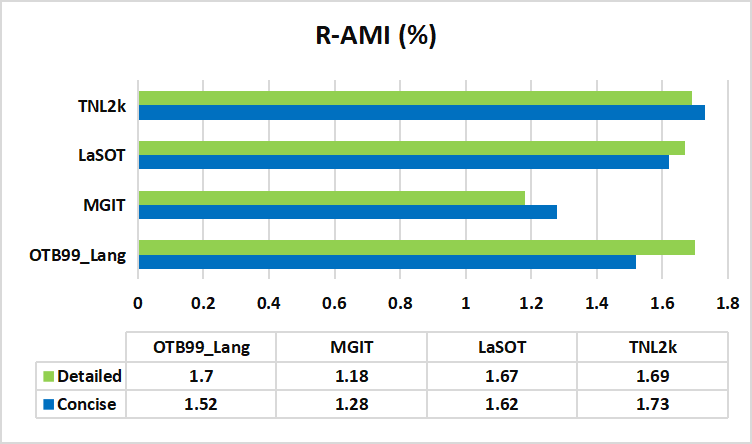}
    \end{minipage}\hfill
    \begin{minipage}{0.48\textwidth}
        \centering
        \includegraphics[width=\textwidth]{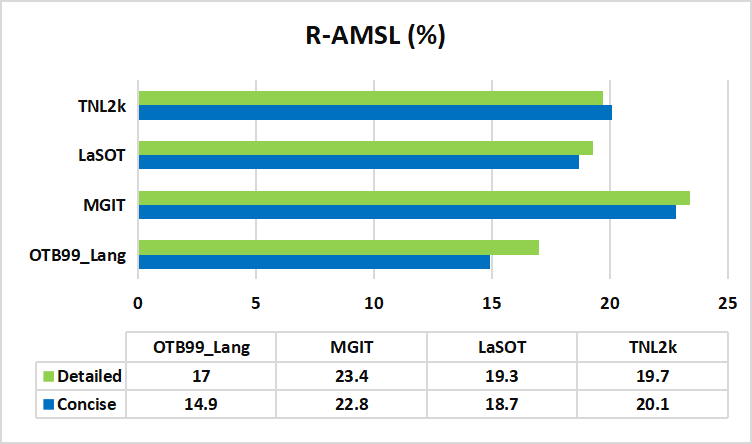}
    \end{minipage}
    \caption{Comparison of robustness on VLT-MI with R-AMI and R-AMSL. The calculation of relative metrics is based on absolute metrics, divided by the sequence length and then averaged.}
    \label{figure}
    \vspace{-16pt}
\end{figure}

We selected four representative datasets—OTB99\_Lang \cite{otb99}, LaSOT \cite{lasot}, TNL2K \cite{tnl2k}, and MGIT \cite{mgit}—to evaluate short-term, long-term, and global instance tracking tasks. For our baseline model, we chose the state-of-the-art VLT tracker, JointNLT \cite{jointnlt}, and evaluated its performance across all four benchmarks. To ensure a fair comparison of tracking performance, we directly tested the model using the official weights, assessing precision (PRE), normalized precision (N-PRE), and success rate (SR) \cite{git}. Additionally, we reported the average multi-modal interaction number (AMI), relative average multi-modal interaction number (R-AMI), average maximum tracking success length (AMSL), and relative average maximum tracking success length (R-AMSL) to measure the tracker's robustness. The results are presented in Table \ref{tracking}, Table \ref{robust}, and Figure \ref{figure}.

\begin{table}[t!]
    \vspace{-10pt}
    \centering
    \fontsize{8pt}{10pt}\selectfont
    \setlength{\tabcolsep}{3.5pt}
    \vspace{-8pt}
    \caption{Comparison of tracking accuracy on VLT-MI. The best two results are highlighted in {\color{red}red} and {\color{blue}blue}.}
    \label{tracking}
    \begin{tabular}{lcccccccccccc}
    \toprule
     \multicolumn{1}{c}{\multirow{2}{*}{JointNLT \cite{jointnlt}}}
      & \multicolumn{3}{c}{OTB99\_Lang \cite{otb99}} & \multicolumn{3}{c}{MGIT \cite{mgit}} &\multicolumn{3}{c}{LaSOT \cite{lasot}} & \multicolumn{3}{c}{TNL2K \cite{tnl2k}} \\ \cmidrule(lr){2-4} \cmidrule(lr){5-7} \cmidrule(lr){8-10} \cmidrule(lr){11-13} 
      & PRE & N-PRE & SR & PRE & N-PRE & SR & PRE & N-PRE & SR & PRE & N-PRE & SR \\
      \midrule
      Traditional              & \color{red}{65.0} & \color{red}{54.1} & \color{red}{50.5} & \color{red}{29.8} & \color{red}{66.4} & \color{red}{47.7} & 54.0 & 55.2 & 53.2 & \color{red}{48.7} & \color{blue}{52.1} & \color{blue}{50.0} \\ 
      Interaction\_Concise     & \color{blue}{64.8} & \color{blue}{53.1} & \color{blue}{49.3} & \color{blue}{29.2} & \color{blue}{63.1} & \color{blue}{48.3} & \color{red}{56.2} & \color{red}{58.2} & \color{red}{56.3} & 48.1 & 52.0 & 49.9 \\
      Interaction\_Detailed    & 63.7 & 52.3 & 48.8 & 29.0 & 62.9 & 48.0 & \color{blue}{55.1} & \color{blue}{57.8} & \color{blue}{56.1} & \color{blue}{48.2} & \color{red}{52.3} & \color{red}{50.2} \\ 
    \bottomrule
    \end{tabular}
    \vspace{-20pt}
\end{table}

\begin{wraptable}{l}{0.5\textwidth}
    \centering
    \vspace{-10pt}
    \caption{Comparison of robustness on VLT-MI. We report AMI and AMSL.}
    \vspace{-8pt}
    \label{robust}
    \fontsize{8pt}{10pt}\selectfont
    \setlength{\tabcolsep}{3.5pt}
    \resizebox{0.5\textwidth}{!}{
    \begin{threeparttable}
    \begin{tabular}{lcccccccc}
    \toprule
     \multirow{2}{*}{JointNLT \cite{jointnlt}} & \multicolumn{2}{c}{OTB99\_Lang \cite{otb99}} & \multicolumn{2}{c}{MGIT \cite{mgit}} \\ \cmidrule(lr){2-3} \cmidrule(lr){4-5} 
     & AMI & AMSL & AMI & AMSL \\
    \midrule
    Interaction\_Concise & 8 & 118 & 229 & 2586 \\
    Interaction\_Detailed & 10 & 150 & 235 & 2661 \\ 
    \midrule
    \multirow{2}{*}{JointNLT \cite{jointnlt}} & \multicolumn{2}{c}{LaSOT \cite{lasot}} & \multicolumn{2}{c}{TNL2K \cite{tnl2k}} \\ \cmidrule(lr){2-3} \cmidrule(lr){4-5} 
    & AMI & AMSL & AMI & AMSL \\
    \midrule
    Interaction\_Concise & 37 & 431 & 13 & 171 \\
    Interaction\_Detailed & 40 & 455 & 13 & 163 \\ 
    \bottomrule
    \end{tabular} 
    \end{threeparttable}}
    \vspace{-20pt}
\end{wraptable}

\textbf{Tracking Accuracy.} 
The original intention of interacting with the tracker is to restore tracking when the tracker fails over an extended period, thereby improving tracking accuracy. However, while tracking accuracy on LaSOT \cite{lasot} meets expectations, performance on several other benchmarks has declined after the interaction. We believe this is because the tracker is not yet fully adapted to the multi-round interaction mode and is overly sensitive to the textual input. This behavior is similar to completing the VLT task by "memorizing the answers."

\textbf{Robustness.} 
We evaluate the robustness of the interactive tracking algorithm from two perspectives: the average number of interactions and the average maximum length of successful tracking sequences. As shown in Table \ref{robust} and Figure \ref{figure}, the number of interactions gradually increases as task difficulty rises and sequence length extends. The number of interactions ranges from 8 to 13 for short-term tracking, to 37 for long-term tracking, and up to 229 for global instance tracking. R-AMI reflects the proportion of interaction frames within the sequence, while R-AMSL represents the percentage of the longest successful tracking subsequence in the sequence. As tracking progresses from short-term to long-term and global instance tracking, the proportion of interaction frames steadily decreases. Among the four benchmarks, MGIT \cite{mgit} exhibits the highest proportion of longest successful tracking subsequences, whereas OTB99\_Lang \cite{otb99} has the lowest, which is consistent with expectations.

\section{Conclusions}

VLT extends the SOT task by introducing a textual modality, which naturally enhances the interactive capabilities between the tracker and humans. In this paper, we present VLT-MI, the first implementation of multi-round, multi-modal interaction within object tracking. Interactions are facilitated through textual updates and target recovery when the tracker repeatedly fails to follow a specific object. We analyze interactive behaviors and robustness, aiming to provide new insights into the advancement of visual language trackers.

From our perspective, human-computer interaction is a critical objective for video language tasks, as demonstrated by VLT. We explore how to integrate human factors into video language tasks to support multi-modal interaction, and we introduce a novel robustness evaluation method from an interaction standpoint. We hope that this work can be extended to more video language tasks, further advancing the development of video language models.

{
\bibliographystyle{plain}
\bibliography{neurips_2024}
}

\end{document}